\documentclass{article}

\usepackage{microtype}
\usepackage{graphicx}
\usepackage{subfigure}

\usepackage{booktabs} % for professional tables
\usepackage{epsfig,amsmath,amssymb,amsfonts,mathrsfs}

\usepackage{hyperref}

\usepackage[accepted]{icml2019_phys4dl}

\icmltitlerunning{Mean field theory of activation functions in Deep Neural Networks}

\begin{document}

\twocolumn[
\icmltitle{Mean Field Theory of Activation Functions in Deep Neural Networks}

\icmlsetsymbol{equal}{*}

\begin{icmlauthorlist}
\icmlauthor{Mirco Milletar\'i}{mic}
\icmlauthor{Thiparat Chotibut}{sutd,chula}
\icmlauthor{Paolo E. Trevisanutto}{ects,tif,bk}
\end{icmlauthorlist}

\icmlaffiliation{mic}{Microsoft, Singapore}
\icmlaffiliation{sutd}{Singapore University of Technology and Design, Singapore}
\icmlaffiliation{chula}{Department of Physics, Faculty of Science, Chulalongkorn University, Bangkok, Thailand}
\icmlaffiliation{ects}{European Center for Theoretical Studies in Nuclear Physics and Related Areas (TN), Italy}
\icmlaffiliation{tif}{Trento Institute for Fundamental Physics and Applications (TN), Italy}
\icmlaffiliation{bk}{Fondazione Bruno Kessler (TN), Italy}

\icmlcorrespondingauthor{Mirco Milletar\'i}{milletari@gmail.com}
\icmlcorrespondingauthor{Thiparat Chotibut}{Thiparat.C@chula.ac.th, thiparatc@gmail.com}

\icmlkeywords{Machine Learning, ICML, activation functions, statistical mechanics}

\vskip 0.3in
]

\printAffiliationsAndNotice{} % otherwise use the standard text.

\begin{abstract}
We present a Statistical Mechanics (SM) model of deep neural networks, connecting the energy-based and the feed forward networks (FFN) approach.
We infer that FFN can be understood as performing three basic steps: {\it encoding}, { \it representation validation} and {\it propagation}. From the meanfield solution of the model, we obtain a set of natural activations -- such as {\it Sigmoid}, $\tanh$ and {\it ReLu} -- together with the state-of-the-art,
{\it Swish}; this represents the expected information propagating through the network and tends to
{\it ReLu} in the limit of zero noise.
We study the  spectrum of the Hessian on an associated classification task, showing that {\it Swish} allows for more consistent performances over a wider range of network architectures.
\end{abstract}

\section{Introduction}

Advances in modern computing hardware and availability of massive datasets have empowered multilayer artificial neural networks, or deep learning (DL), with unprecedented capabilities for image and speech recognition tasks. Despite these empirical success, theoretical understanding of why and when multilayer neural networks perform well lags far behind~\cite{mezard1}.
Only recently, theoretical efforts in this direction have been intensively reported. For example, recent works shed light on how FFN attains its expressive power~\cite{Poggio2017, LinTegmark2017, Raghu2017, Poole2016}, what contributes to its generalizability~\cite{Zhang2017,Dinh2017}, and how myriad parameters in the network affect the geometry of the loss function ~\cite{dauphin, ch, penn1, penn2}. Taken together, these results have paved the way for a systematic design of robust and explainable FFNs.

Using modern optimization and regularization techniques such as dropout~\cite{srivastava2014}, non-linear activation functions, and complex network architectures~\cite{Krizhevsky}, to name a few, the FFN can be efficiently trained on large-scale datasets such as ImageNet or CIFAR to achieve low training and generalization errors.
While these engineering feats improve the performance of FFN, a clear design principle is still lacking, leading to an unsystematic growth of model complexity.

To assist future systematic studies and construction of FFNs, we propose a theoretical framework based on the tool-set of SM. It allows for the definition of an energy based model in which a hidden unit is regarded as a communication channel, first encoding and then {\it transmitting} the result of its computation through a gate with a specific transmission probability;  the latter is obtained via the maximum entropy principle~\cite{zecchina, jaynes} under the biologically inspired constraint that a neuron responds by firing (transmit its signals) or not with a certain probability.
By interpreting a hidden unit as a communication channel, its activation takes the form of the expected (mean) signal transmission; remarkably, this activation corresponds to {\it Swish}, obtained in ~\cite{elfwig, prajit} through an extensive search algorithm trained with reinforcement learning and shown to best perform on the CIFAR and ImageNet datasets among all candidate activations. Although some activations may perform better for a specific datasets, they generally fail to generalize. Finally, the {\it ReLu} activation arises as a limiting, noiseless case of {\it Swish}. To the best of our knowledge, this provides the first  derivation of {\it ReLu}, typically introduced heuristically to facilitate optimization. Despite restricting our analysis to pure FFNs, most of our conclusions carry on to Convolutional and Recurrent networks.

\section{Motivation and Model} \label{sec:model}

\subsection{General Setup} \label{sub:setup}

A standard task in supervised learning is to determine an input/output relation between a set of $m$ features and the observed labeled outcomes. Let us denote with $x^{\mu}_i$ the input vector, where $i \in [1, n]$ denotes features, and $\mu \in [1,m]$ denotes examples; we also denote the output as $y^{\mu}_k$, where $k$ is the number of classes. Quite generally, the probability of measuring the output $\mathbf{y}$ given the input can be written as $P(\mathbf{y}) = \int d\mathbf{x} \, P(\mathbf{y} | \mathbf{x}) \, P(\mathbf{x})  =  \int d \hat{\mathbf{y}} \, P(\mathbf{y} | \hat{\mathbf{y}} ) \, P(\hat{\mathbf{y}})$ where we have used the chain rule~\eqref{a:loss} and introduced the output layer ($\hat{ \mathbf{y}}$). Once $P(\hat{\mathbf{y}})$ has been learned, one can obtain the loss function by taking the log-likelihood $\mathscr{L} = - \log P(\mathbf{y})$; for example, in a binary classification problem, $P(\hat{\mathbf{y}}[\boldsymbol{\theta}])$ is Bernoulli and parametrized by $\boldsymbol{\theta_l} = \{\mathbf{W}_l,\mathbf{ b}_l \}$, being the weights and biases of the neural network, with $l \in [1, L]$ the number of hidden layers. In this case, the loss function is the binary cross-entropy, but other statistical assumptions lead to different Losses; see~\eqref{a:loss} for a meanfield derivation of the cross-entropy loss. To motivate the model, in the next section we begin by discussing a (physical) dimensional inconsistency in the standard formulation of {\it forward propagation} in FFNs, and how this can be reconciled within our framework.
\subsection{Motivations} \label{sub:mot}
Motivated by neurobiology and in analogy with the communication channel scheme of information theory~\cite{mckay, jaynes}, we regard the input vector $x^{\mu}_i$ as the information source, while the units constitute the encoders. Quite generally, the encoders can either build a lower (compression) or higher dimensional (redundant) representation of the input data by means of a properly defined transition function. In a FFN, the former corresponds to a compression layer while the latter to an expansion layer. If the encoded information constitutes an informative representation of the output signal (given the input), it is passed  over to the next layer for further processing until the output layer is reached. In the biological neuron, this last step is accomplished by the synaptic bouton, that releases information whether or not the input exceeds a local bias $b_i$. Both in the brain and in electronic devices, the input is often conveyed in the form of an electric signal, with the electron charge being the basic unit of information, the signal has dimension (units) of $Coulomb$ in SI. For an image, the information is the brightness level of each pixel and more in general, information has units of bits. Clearly, a linear combination of the input signals needs to preserve dimensions: $h^{\mu}_i = \sum_{j=1}^{n} \, w_{ij} x^{\mu}_j + b_i$, where $i \in [1, n_1]$ indices the receiving units in the first layer and the weight matrix $ w_{ij}$ is the coefficient of the linear transformation and it is dimensionless. For noiseless systems, the input is transmitted i.f.f. the overall signal exceeds the bias $b_i$. However, in the presence of noise, the signal can be transmitted with a certain probability even below this threshold; in the biological neuron, the variance in the number of discharged vesicles in the synaptic bouton and the number of neurotransmitters in each vesicle is responsible for such noisy dynamics~\cite{amit1}. Let us now consider the {\it Sigmoid} non-linearity $\sigma(\beta \, \mathbf{h} )$, where $\beta$  has inverse dimension of $\mathbf{h}$;  $\sigma$ expresses the probability of the binary unit to be active and can be seen as an approximation of the biological neuron's firing probability~\cite{amit1}. Being a {\it distribution} defined in $[0,1]$, $\sigma$ is intrinsically {\it dimensionless}. The parameter $\beta$ defines the spread of the distribution and one typically sets it to $1$ or reabsorb it inside the weights and bias~\cite{zecchina}; here we keep it general for reasons that will be clear later. Defining $\mathbf{a} = \sigma(\beta \, \mathbf{h} )$ as the input of the next layer, we can immediately see the dimensional mismatch: a linear combination of $\mathbf{a}$ is dimensionless and when passed through the new non-linearity, $\sigma(\beta \, \mathbf{a} )$ necessarily becomes dimensionful. In the next section we show how this simple fact relates to gradients vanishing during back propagation. From a conceptual point of view, one is transmitting the expectation value of the transmission gate (synapse) rather than the processed signal. This problem persists with the $\tanh$ activation, but is resolved when using {\it ReLu}, that correctly transmits the expected value of the information.

\subsection{Statistical Mechanics of Feed Forward networks}

A prototypical SM formulation of Neural Networks is the {\it inverse} Ising model~\cite{zecchina}, or Boltzmann machine, where one infers the values of the couplings and external fields given a spin configuration.  While the Boltzmann machine is an energy-based model, a standard FFN is not commonly regarded as such. Here, we propose to bridge on these two formulations using the maximum entropy principle~\cite{zecchina, roberto, mckay, jaynes} to obtain the least biased representation of hidden neurons. Starting from the input layer, each unit takes the same input vector $\mathbf{x}$ and outputs a new variable $\mathbf{h}$. We now regard $h_i$ as a coarse grained field coupling to the synaptic gate variable $s_i$. The feedforward nature of coarse graining (a directed graph), stems from its irreversibility and endorses the forward pass with a semi-group structure. Considering the first layer, we need to evaluate the probability associated to  $\mathbf{h}$, $P(\mathbf{h}) =  \int d \mathbf{x} \, Q(\mathbf{h} | \mathbf{x} ) \, P(\mathbf{x})$, where  $Q(\mathbf{h} | \mathbf{x} )$ is the transition function modeling the encoder. In DL, the latter is fixed by the forward pass, while the input data are drawn from some unknown distribution $P(\mathbf{x})$. We can follow two different paths: fix the value of $\mathbf{x}$ on the observed sequence, or assume the form of the distribution from which $\mathbf{x}$ has been sampled from; here we choose the former option. Consider then the empirical estimator $P(\mathbf{x}) =  \prod_{\mu=1}^m  \delta( \mathbf{x} - \mathbf{x}^{\mu} )$, where the Dirac-delta function  fixes the input on the observed sequence. As for the transition function, it enforces the information processing performed by the ``soma'' of the artificial neuron; in DL, this consists of creating a linear combination of the input information. Information conservation enforces an additional constrain on the weights, $\sum_i \, w_{ij}^{[l]} =1 \, \forall \, l,j$, formally equivalent to the conservation of charge in physics
\begin{align} \label{eq:ps2}
P(\mathbf{h}) &= \int {d\mathbf{x}} \, \Gamma^{[1]} \, \delta\left( \mathbf{h} -  \mathbf{w}^{T} \, \mathbf{x} - \mathbf{b} \right) \prod_{\mu =1}^m  \delta ( \mathbf{x} - \mathbf{x}^{\mu} ) \\ \nonumber
\Gamma^{[1]} &= \prod_{j=1}^n \delta\left( \sum_{i=1}^{n_1} w_{ij}^{[1]} -1 \right).
\end{align}
Eq.~\eqref{eq:ps2} is akin to the real space block-spin renormalization developed by Kadanoff, reformulated in the more general language of probability theory~\cite{roberto, ma, cassandro}, albeit without proper rescaling. Alternatively, Eq.~\eqref{eq:ps2} can be seen as a functional ``change of variables'', with $\mathbf{h}$ a cavity field. The relation between DL and the renormalization group was previously observed in the context of restricted Boltzmann machine (RBM)~\cite{mehta}. Once $\mathbf{h}$ has been computed, the information passes through the ``synaptic'' gate, and transmitted with a certain probability if it exceeds the threshold $b_i$. The core task at hand is to determine the state of the gate (open/close). The absence of lateral connections in FFNs means that each synaptic gate is only influenced by its receptive signal. Given a statistical ensemble of hidden binary gates, the least unbiased distribution  can be obtained by maximizing its entropy, subject to constraints imposed by the conserved quantities, typically the first and second moments of the data~\cite{zecchina, mckay}. However, in the absence of lateral connections, the entropy functional of the hidden gates $s_i$ does not account for the second moment and it reads:
\begin{align} \label{eq:entropy2}
\mathscr{F} &= - \sum_{\mathbf{s}} P(\mathbf{s}) \, \log P(\mathbf{s}) + \eta \left( \sum_{\mathbf{s}} P(\mathbf{s}) -1\right) \\ \nonumber
&+ \sum_i \, \lambda_i \left( m_i -  \sum_{\mathbf{s}} \, s_i \, P(\mathbf{s}) \right),
 \end{align}
where $\lambda_i$ are Lagrange multipliers chosen to reproduce the first moment of the data,  while $\eta$ enforces normalization. Functionally varying with respect to $P(\mathbf{s)}$ and solving for the Lagrange multipliers, one  obtains~\cite{roberto} the conditial probability:
\begin{equation} \label{eq:entropy4}
P(\mathbf{s} | \mathbf{h}) = \frac{e^{ \sum_i \beta_i  \, s_i \, h_i }}{\prod_i \, \left( 1+ e^{\beta_i \, h_i} \right)}
\end{equation}
where $\beta_i$ encodes the noise.  In physics, $\beta_i$ is the inverse temperature in units of Boltzmann's constant and in equilibrium it is the same for each $s_i$; however, here the network is only in a local equilibrium as the units are not allowed to exchange ``energy'' (information) among themselves due to the lack of pairwise interactions. We have introduced $P(\mathbf{s} | \mathbf{h})$ to denote the conditional probability of $\mathbf{s}$ given the signal $\mathbf{h}$ --  and the partition function $Z$. Finally, given the distribution of the coarse grained inputs and the conditional probability of the channels $P(\mathbf{s}|\mathbf{h})$, one is left evaluating the channel transmission probability
\begin{equation} \label{eq:ps3}
P(\mathbf{s}) = \int d \mathbf{h} \, P(\mathbf{s} | \mathbf{h}) \, P(\mathbf{h}) =  \frac{e^{-\sum_i \, \beta_i \mathscr{H}_i[s, x^{\mu}]}}{m^{-1}\prod_{\mu,i} Z_{i, \mu}}
\end{equation}
where $\hat{h}^{\mu}_i = \sum_{j} w^{[1]}_{ij} \, x^{\mu}_j + b_i $, \,  $Z_{i, \mu}= 1+ e^{\beta_i \, h^{\mu}_i}$ is the partition function per example/index and we have identified the coarse grained Hamiltonian $\mathscr{H}_i = -  \, \sum_{j=1}^n \, s_i \, w^{[1]}_{ij} x^{\mu}_j -  \, s_i \, b_i$. Eq.~\eqref{eq:ps3} is the starting point of an {\it energy based model}, where it is the coarse grained probability $P(\mathbf{s})$ that propagates through the network (see e.g.~\cite{connie} ) as opposed to signals in a FFN; $\mathscr{H}$ has the form of a RBM with binary hidden units. The expected value of the channel transmission, $\langle s_i \rangle$ is~\cite{hertz} the logistic function, for each empirical realization $\mu$ and unit $i$:
\begin{equation} \label{eq:ps4}
\langle s_i \rangle = {\beta_i}^{-1} \partial_{b_i} \log Z_{i, \mu} =  \sigma(\beta_i \, \hat{h}^{\mu}_i).
\end{equation}
This is the known mean-field solution of the Ising model; here it is exact due to the lack of lateral connections. We stress that this is not the coarse grained input signal {\it transmitted} by the channel; it solely determines the expectation of channel transmissions {\it given} $h_i^{\mu}$. To ensure dimensional consistency across hidden layers, the output of each unit must have the same dimension as its input. Therefore, the correct quantity to consider is the {\it expectation value} of the output $j_i = h_i \, s_i $.  This can be  obtained by summing over all gate states, or by using the partition function of Eq.~\eqref{eq:ps3}
\begin{equation} \label{eq:ps5}
\langle j_i \rangle = \langle \hat{h}_i^{\mu} \, s_i \rangle_s = \partial_{\beta_i} \log{Z_{i,\mu}} = \hat{h}^{\mu}_i \, \sigma(\beta_i \, \hat{h}^{\mu}_i),
\end{equation}
that agrees with the definition of the energy flux in SM~\cite{bellac}. Note that contrary to Eq.~\eqref{eq:ps4}, the noise parameters $\beta_i$ cannot be rescaled away. This function was recently obtained in ~\cite{prajit} ({\it Swish}), through an extensive search algorithm trained with reinforcement learning. In their extensive search, the authors found that activations of the form $a = x f(x)$  better performed on several benchmark datasets. A theoretical study of the performance of {\it Swish} from the point of view of information propagation has been proposed in ~\cite{soufiane}.  Our model naturally complements these studies by identifying {\it Swish} with the expectation value of the coarse grained input transmitted by each unit. In the noiseless limit:
\begin{equation} \label{eq:ps6}
\lim_{\beta_i \to \infty } \langle j_i \rangle =  \hat{h}^{\mu}_i \, \theta(\hat{h}^{\mu}_i)  \equiv   \rm{max} \left\{ \hat{h}^{\mu}_i, 0\right\} = ReLu(\hat{h}^{\mu}_i),
\end{equation}
where $\theta(.)$ is the Heaviside step function and in the last equality we have identified {\it ReLu}. Although this limit was already noted heuristically in \cite{prajit}, to the best of our knowledge the one presented here is the first consistent derivation and theoretical explanation of {\it ReLu}. {\it ReLu} emerges as the noiseless limit of the mean transmitted information across the units; as such, it is not affected by the dimensional mismatch, see Sec.~\eqref{sub:mot}. We would like to stress that both {\it Swish} and {\it ReLu} pass on the expected value of their computation; the latter does it with probability one {\it only} if the input signal exceeds a threshold, while the former can pass a signal lower than the threshold with a finite probability. A positive signal means that a certain combination of the inputs should be strengthen while a negative one means that it should be weakened, i.e. unlearned; the latter option is absent when using {\it ReLu}. In the opposite limit $\beta \ll 1$, {\it Swish} becomes linear; therefore, we can consider linear networks as a noisless limit of non linear ones. The effect of noise in FFNs was recently considered in ~\cite{pratik}, where an optimization scheme based on the maximum entropy principle was proposed.

\subsection{On Back-propagation} \label{sub:back}
At the heart of any FFN model is the back propagation algorithm~\cite{hertz, bishop}. Consider the output layer ``L'', then the gradient of the weights is
\begin{equation} \label{eq:gradsL}
\frac{\partial \mathscr{L}}{\partial \mathbf{w}_L } =  \frac{1}{m} \sum_{\mu=1}^m \, \left[ \mathbf{e}^{\mu} \, \mathbf{g}(\hat{\mathbf{h}}^{\mu} _L) \right]  \mathbf{a}_{L-1}
 \end{equation}
where $\mathbf{e}^{\mu}\propto y^{\mu} - \hat{y}^{\mu}$ is the residual error and $\hat{\mathbf{g}}(\mathbf{h}^{\mu}_l) =  \boldsymbol{\sigma}(\boldsymbol{\beta}_l \, \hat{\mathbf{h}}^{\mu}_l)[ 1 + \hat{\mathbf{h}}^{\mu}_l \boldsymbol{\beta}_l \, \boldsymbol{\sigma}(-\boldsymbol{\beta}_l \, \hat{\mathbf{h}}^{\mu}_l) ] $. In the optimization phase we look for stationary points defined by $\partial \mathscr{L}/\partial \boldsymbol{\theta}^{\alpha}_l=0$, where $\boldsymbol{\theta}^{\alpha}_l = \left\{ \mathbf{w}_l, \mathbf{b}_l, \boldsymbol{\beta}_l \right\} $. For a linear network $\mathbf{g}(\hat{\mathbf{h}}^{\mu} _L)=const$ and the latter condition is  strictly satisfied if $ \mathbf{e}^{\mu}=0$. However, in a non linear network we can also have $g(\hat{\mathbf{h}}^{\mu})=0$. In principle, there may be situations in which $\mathbf{e}^{\mu}$ is far from zero but $g(\hat{\mathbf{h}}^{\mu}) \simeq 0$, in which case learning will not be effective. For a $\sigma$ activation, $\mathbf{g}(\hat{\mathbf{h}}^{\mu}_l) = \boldsymbol{\sigma}(\boldsymbol{\beta}_l \, \hat{\mathbf{h}}^{\mu}_l)\boldsymbol{\sigma}(-\boldsymbol{\beta}_l \, \hat{\mathbf{h}}^{\mu}_l)$ \footnote{in physics, the phase space factor~\cite{roberto}.}%, describing the transition probability from an occupied to an empty state
~\cite{tishby1, tishby2} show that there are two distinct phases of learning: empirical error minimization (the residual) and representation learning. We can identify the latter with the task of optimizing $g(\hat{\mathbf{h}}^{\mu})$. When $\hat{h}_i \gg 1/\beta_i$, i.e. when the signal greatly exceeds the noise, then $\sigma( \beta_i \,\hat{ h}_i) \equiv P(s_i=1| \bar{x}_i)  \simeq 1$ ($\bar{x}_i$ is the unit's input) and the back propagated signal is small, being proportional to $g(\hat{h} ) \simeq 0$. We then have the paradoxical situation in which, although the lower dimensional representation of the information is considered to be relevant, learning is likely to be inefficient.

Consider now the same limiting behavior for {\it Swish}. If $\hat{h}_i \gg 1/\beta_i$, $g(\hat{h}^{\mu}_i)\simeq 1$, i.e. the representation learning phase is completed and learning moves towards minimising the empirical error, see fig.~(\ref{fig:index}.c). In the opposite limit, the signal is much smaller than the noise and learning is impossible. Finally, in the noiseless case, one obtains the {\it ReLu} solution $g(\hat{h}^{\mu}_i) = (1, 0) $, for $\hat{h}_i $ respectively greater or smaller than zero. This corresponds to a network in which representation unlearning is impossible. The fact that {\it Swish} can be negative for $\sigma(\hat{h})/\beta <\hat{h}<0$ allows to avoid plateaus surrounding local minima and saddle-points. In~\cite{dauphin}, it was proposed that loss plateaus of zero (or small) curvatures are responsible for preventing convergence of gradient descent.
\section{Numerical Analysis}\label{sec:numerical}
\begin{figure*}[t!]
\centering
	\begin{subfigure}[]{ }
	\includegraphics[width = 0.48\textwidth]{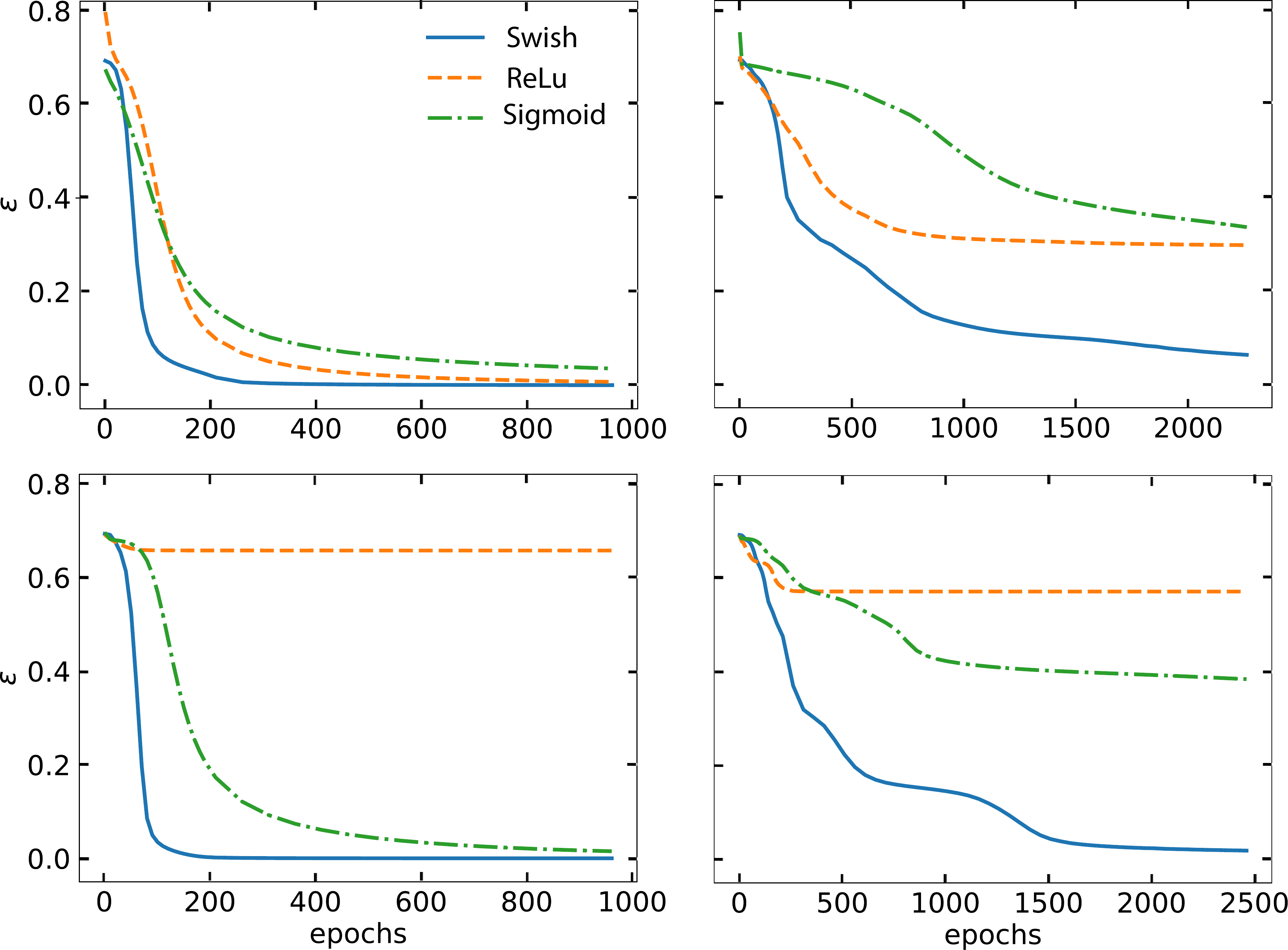}
	\label{fig:cost1}
  \end{subfigure}
 	%~
	\begin{subfigure}[]{}
 		\includegraphics[width = 0.48\textwidth]{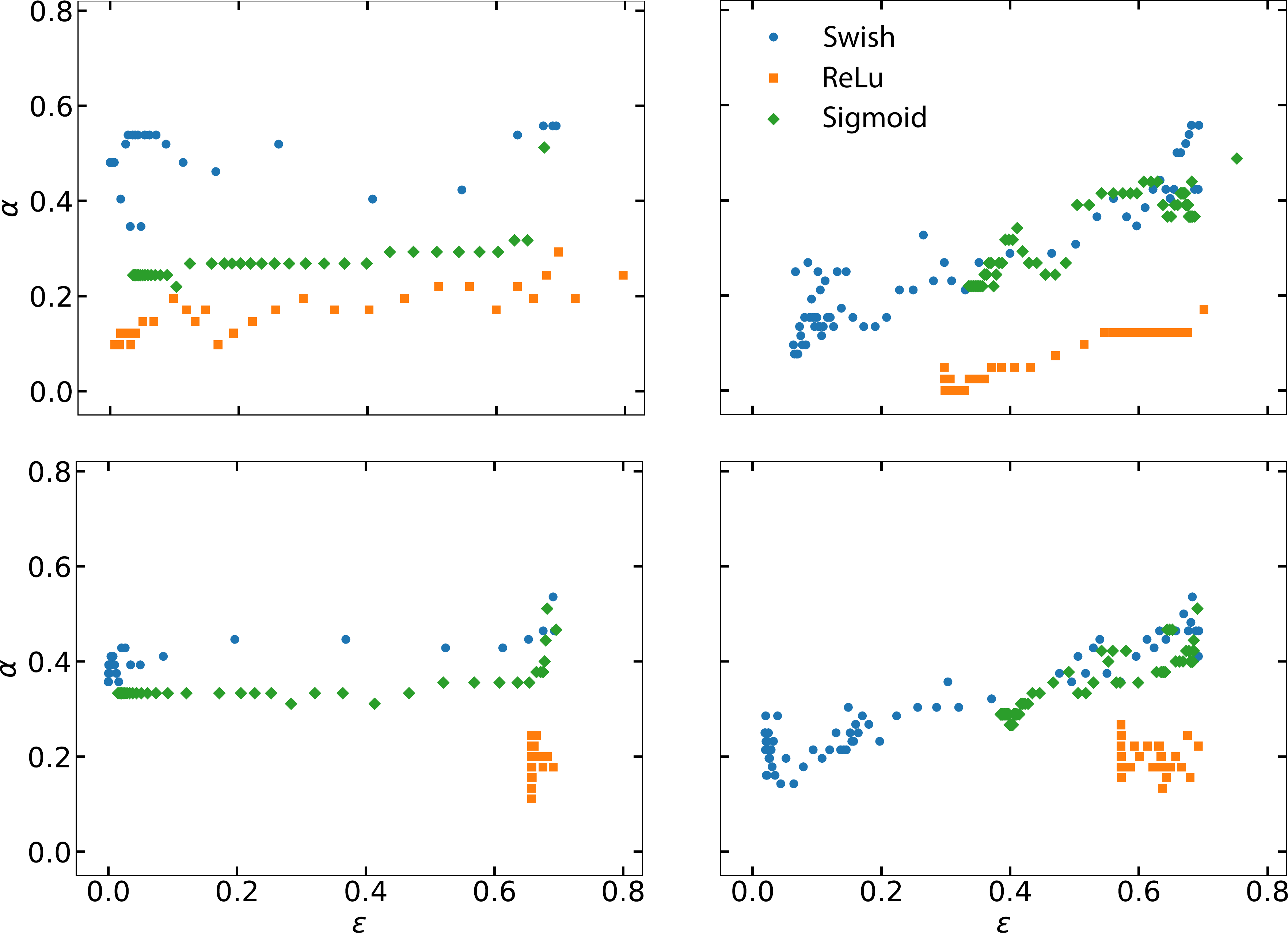}
		\label{fig:index1}
	\end{subfigure}
  ~
  \begin{subfigure}[]{}
	\includegraphics[width= 0.48\textwidth]{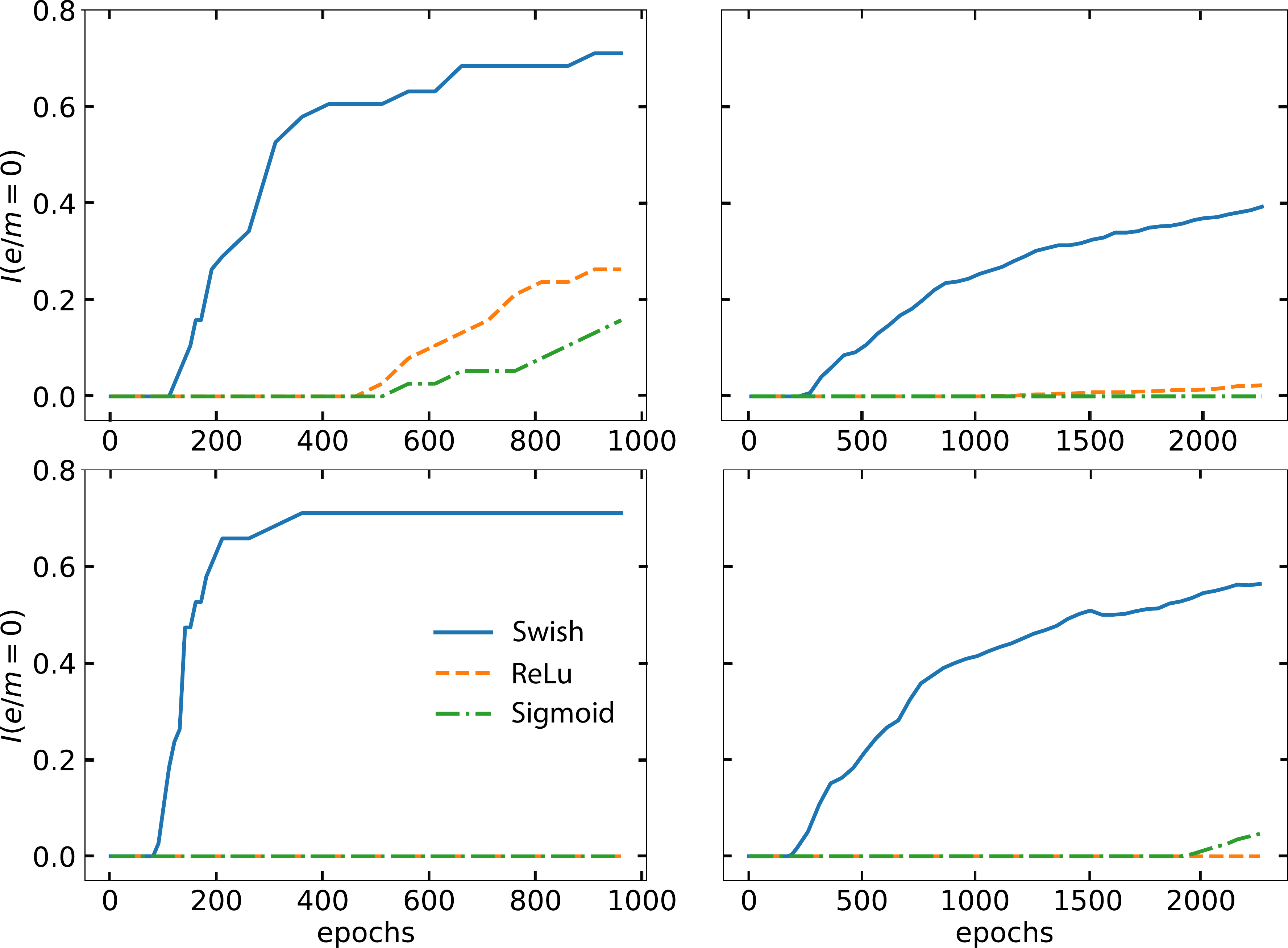}
	\label{fig:res}
	\end{subfigure}
  \begin{subfigure}[]{}
	\includegraphics[width= 0.48\textwidth]{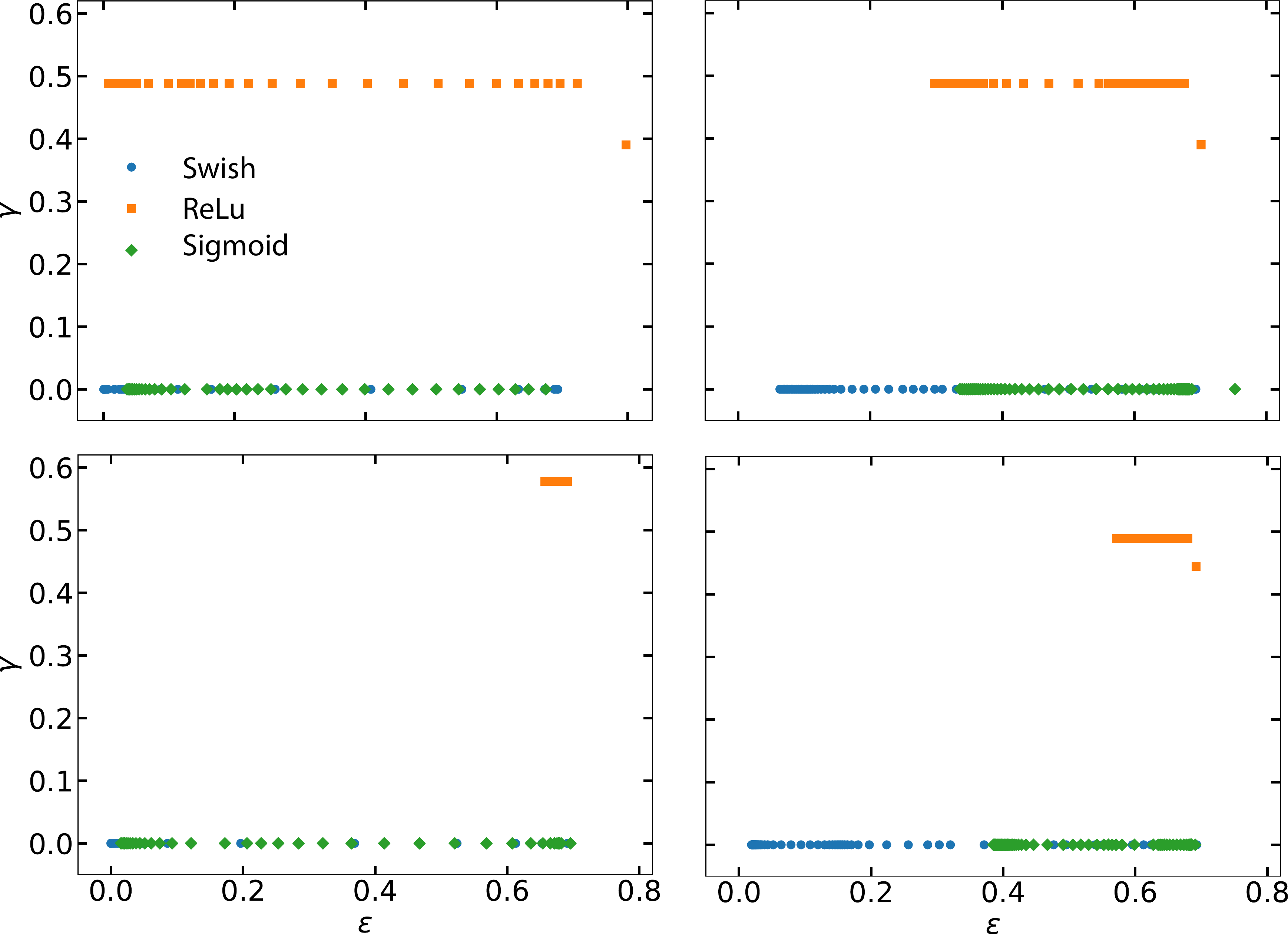}
	\label{fig:gamma}
	\end{subfigure}

	\caption{ \label{fig:index} (a) Loss functions for linear (left) and non-linear (right) binary classification. (Top) 10 units hidden layer (Bottom) two hidden layer of 8 and 2 units respectively. All cases have a {\it Sigmoid} activation in the last layer and a learning rate of 0.01. (b) $\alpha$ index  v.s. energy for the single layer network with linear (left) and non-linear (right) decision boundary. (c)  Fraction of zero residuals as a function of training epochs for the single 8-2 layer network with linear (top/bottom left) and non-linear (top/bottom right) decision boundary. (d) Fraction of zero eigenvalues. (Top left/right) linear/non-linear dataset with a single, 10 units hidden layer. (Bottom left/right)  linear/non-linear dataset with two hidden layers of 8 and 2 units respectively. Curves are evaluated with the same number of epochs within a single batch training. }
\end{figure*}
A thorough benchmark of {\it Swish} versus other activations was presented in ~\cite{prajit}, where it was shown that {\it Swish} outperformed all other choices on image classification tasks. Hereafter,  we consider both artificial and experimental datasets. We considered two binary classification tasks: a linear and a non-linear one, each trained with one and two hidden layers. For the linear task with a single, 10 units layer, all three activations attain full train/test accuracy but {\it Swish} is the fastest converging.  For the non-linear task, {\it ReLu} quickly converges to a suboptimal plateau. To obtain a better understanding, we have evaluated two different indices: the fraction of negative eigenvalues of the Hessian --$\alpha$-- and the fraction of zero eigenvalues -- $\gamma$ --. The former measures the ratio of descent to ascent directions on the energy landscape; when $\alpha$ is large, gradient descent can quickly escape a critical point due to the existence of multiple unstable directions. However, when a critical point exhibits multiple near-zero eigenvalues, roughly captured by $\gamma$, gradient descent will slowly decrease the training loss.
In general, we find that for {\it ReLu} networks $\gamma \neq 0$ across all the training phase, while this is typically not the case for both {\it Swish} and {\it Sigmoid}-networks. Taking the two layer case as a representative example, we show that {\it ReLu} networks are sensitive to fine tuning of the model: choosing a $10-2$ or a $8-5$ configuration over the $8-2$ considered here, greatly improves learning. In stark contrast, {\it Swish} networks exhibit consistent performance over a wider choice of architecture/learning parameters. Although the performance impact might be fairly small for small networks, it certainly plays an important role for larger ones, as discussed in~\cite{prajit}. Looking at the fraction of residuals $e^{\mu} \simeq (\hat{y}^{\mu} - y^{\mu})/m$ closer to zero we found, surprisingly, that {\it Swish} greatly outperforms the other activations in minimizing the empirical error, see~fig.~(\ref{fig:index}.c).
In addition, we find that the eigenvalue distribution obtained with {\it ReLu}  {\it shrinks} with increasing training epochs, giving rise to the singular distribution reported in ~\cite{penn1, levent}. On the other hand, {\it Swish} shows a significantly wider spread in the eigenvalue distribution during training, with larger $\alpha$ index facilitating downhill descent. At the end of training, the fraction of negative eigenvalues is negligible, comparing to those from {\it ReLu} and {\it Sigmoid}, indicating that a local minima is reached. The results of fig.~\eqref{fig:index} have been obtained using Adam for batch training in order to distinguish the effect of noise introduced by {\it Swish} from the one due to random minibatches. Also notice that the two dataset are fairly small: the linearly separable one contains $51$ examples while the non-linear one has $863$ examples.

To verify that the observed effects are not specific to small datasets, we have performed the same analysis on the MNIST dataset, comprising $1797$ examples and obtained a qualitatively similar result, see fig.\eqref{fig:mnist}.
\begin{figure*}[t!]
\centering
\begin{subfigure}[]{}
\includegraphics[width= 0.48\textwidth]{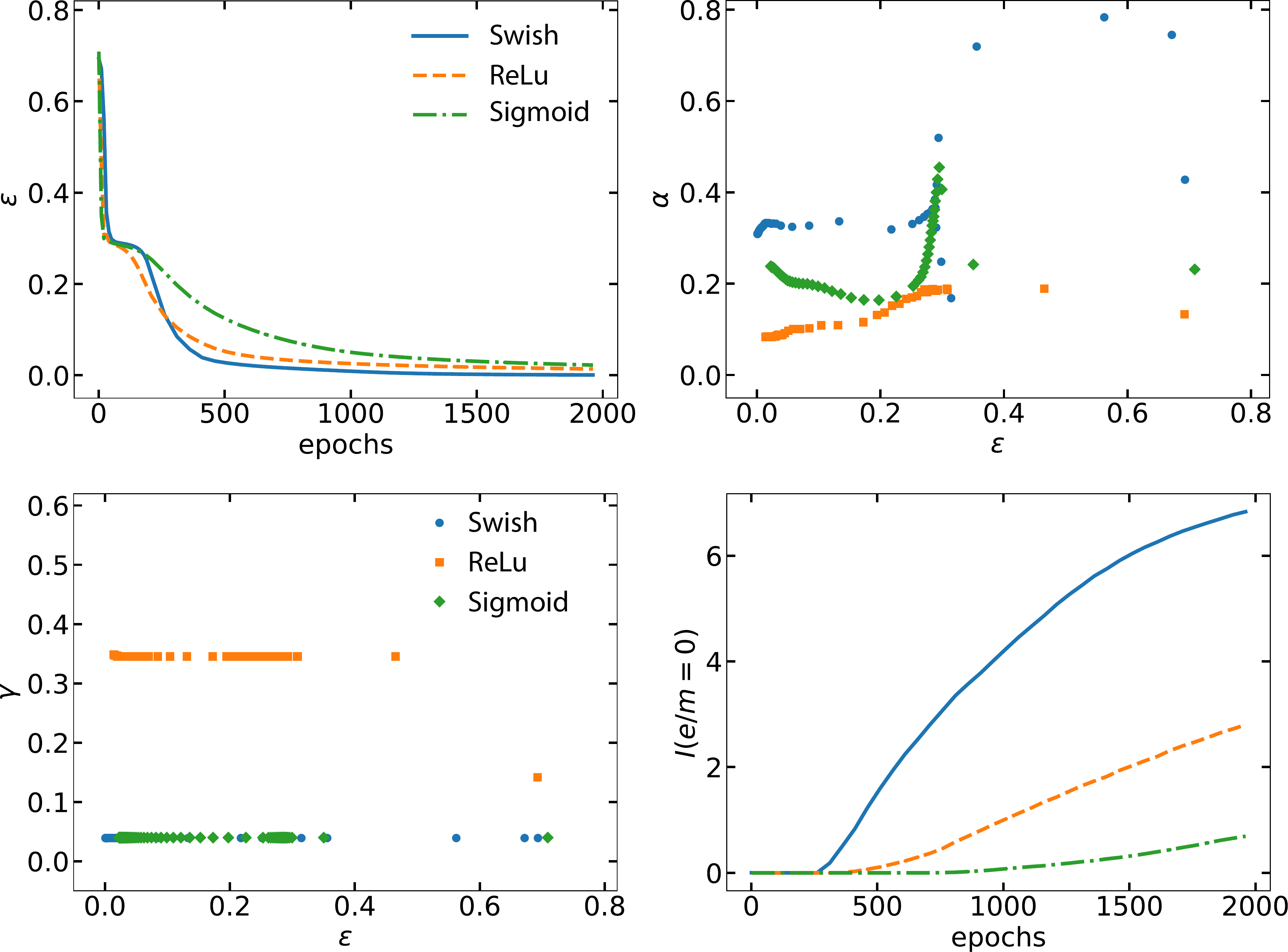}
\label{fig:mnist}
\end{subfigure}
\begin{subfigure}[]{}
\includegraphics[width= 0.5\textwidth]{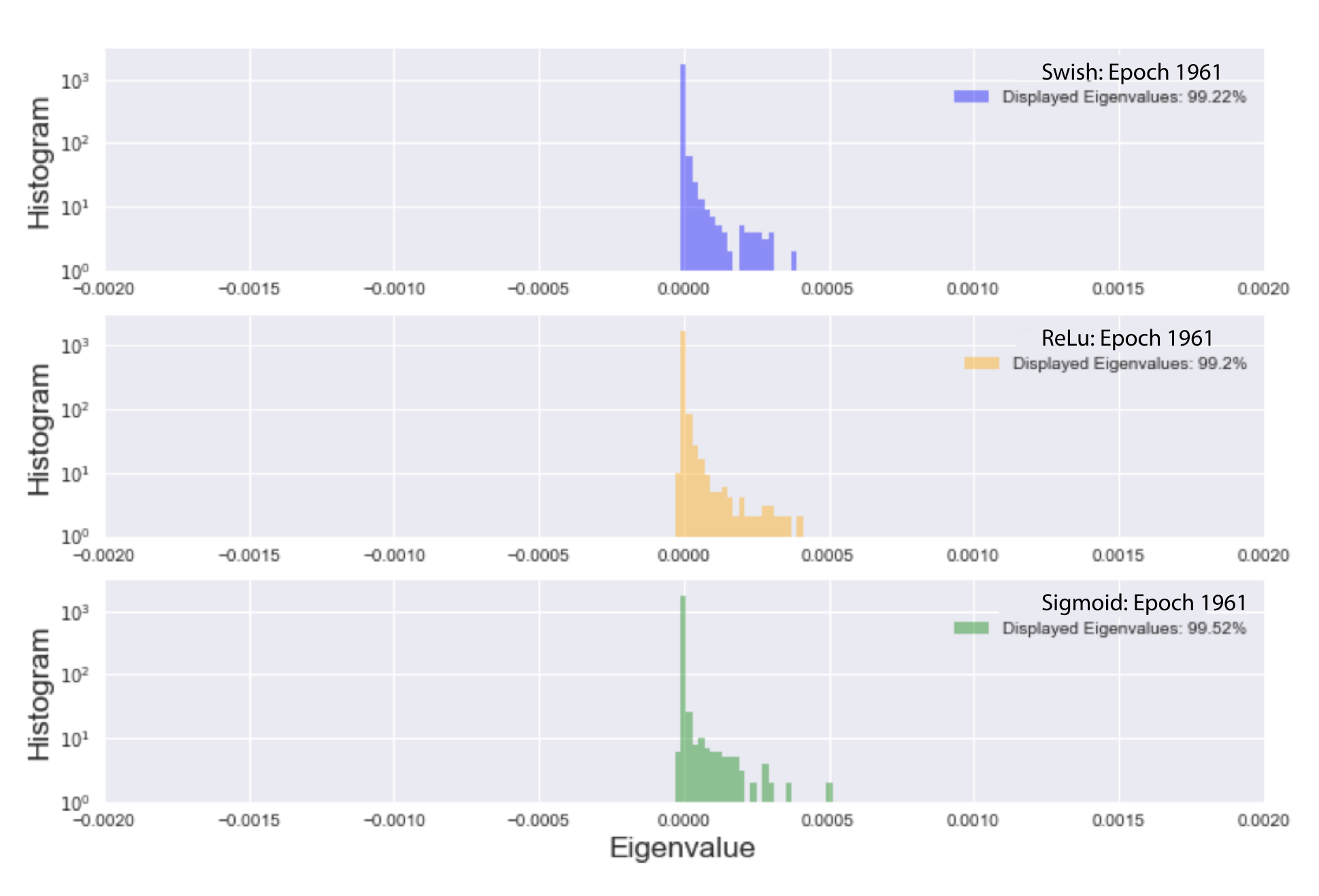}
\end{subfigure}
\caption{ \label{fig:mnist} Single hidden layer network (25 units) trained on the MNIST dataset. (top left) Loss functions, (top right) $\alpha$ index, (bottom left) Fraction of zero eigenvalues $\gamma$ and (bottom right) fraction of zero residuals. (b) eigenvalue distribution at the end of training}
\end{figure*}
Further results and analysis,  can be found in the~\href{https://github.com/WessZumino/meanfield-theory-of-activation-functions}{git repository}.

\section{Conclusion and perspectives}
In this work, we have introduced an energy-based framework that allows systematic construction of FFNs by providing a coherent interpretation of the computational process of each and every hidden unit. The framework overcomes the dimensional mismatch stemming from the use of heuristic activations. Enforcing dimensional consistency naturally leads to activations propagating the mean expectation value of the processed signals, providing a theoretical justification of the \textit{Swish} activation found in~\cite{prajit}. We also demonstrate the superiority of {\it Swish} through numerical experiments that reveal the geometry of its loss manifold that facilitates gradient descent optimization. We hope SM methods can complement standard views of FFNs, e.g. help explore the scaling relation between the {\it Swish} Hessian's eigenvalue distributions and the hyperparameters, similar to the analysis of~\cite{penn1} for standard activations.

\section{Acknowledgements}

This work was initiated when M.M. was at Bambu. M.M. would like to thank Ned Phillips, Bill Phillips, Sarah Chan and Roberto Raimondi for support and discussions. A special thanks to  F\'abio Hip\'olito and Aki Ranin for carefully reading and commenting the manuscript. T.C. would like to thank Shaowei Lin for useful discussions and for financial support from the SUTD-ZJU collaboration research grant ZJURP1600103. P.E.T would like to thank M. D. Costa for help with HPC. Some of the calculations were carried out at the HPC facilities of the NUS Centre for Advanced 2D materials. Finally, M.M. And T.C. would like to thank the AI Saturday meetup initiative organised by Nurture.ai, that sparked some of the ideas presented in this work.

\appendix
\section{Cross Entropy Loss} \label{a:loss}
In this appendix we present a derivation of the cross entropy loss function using large deviation methods and mean field theory~\cite{mezard}. Other loss functions can be obtained following the same method while changing some of the statistical assumptions on the label correlations.

In sect.~\eqref{sub:setup} we have introduced the supervised learning task in terms of the probability of the outcome $y$ given the input $x$; this can be related to the network output by means of the chain rule:
\begin{align} \nonumber
P(\mathbf{y}) &= \int d\mathbf{x} \, P(\mathbf{y} | \mathbf{x}) \, P(\mathbf{x}) = \int d \hat{\mathbf{y}} \, P(\mathbf{y} | \hat{\mathbf{y}} ) \int d\mathbf{j}  P(\hat{\mathbf{y}}|\mathbf{j}) \\
&\times \int d\mathbf{x} \, P(\mathbf{j} | \mathbf{x}) \, P(\mathbf{x})
= \int d \hat{\mathbf{y}} \, P(\mathbf{y} | \hat{\mathbf{y}} ) \, P(\hat{\mathbf{y}}),
\label{eqa:chain}
\end{align}
where $P(\hat{\mathbf{y}})$ is the output probability of the network and $ P(\mathbf{j} | \mathbf{x})$ is the transition probability between the input and hidden layer; additional hidden layers can be added by further use of the chain rule. The accuracy of a classification task -- the number of times the network prediction $\hat{y}$ equals the provided solution $y$ -- corresponds to the conditional probability $P(y|\hat{y}) =  \mathbb{I}(y = \hat{y})$. As $\hat{y}$ is a random variable, we need to find the probability of obtaining the true value $y$ in a series of ``experiments'' in which $\hat{y}$  has a certain probability to be equal to $y$. According to Eq.~\eqref{eqa:chain}, and using the standard relation relating the Loss function to the log-probability, we evaluate
 \begin{align} \label{eq:cl1}
-\log P(\mathbf{y}) = - \log \int d \hat{\mathbf{y}}\, P(\mathbf{y} | \hat{\mathbf{y}}) \, P(\hat{\mathbf{y}}).
 \end{align}
From a physics perspective, the above expression corresponds to an {\it annealed } average~\cite{parisi2, giardina} and we will discuss its meaning at the end of the calculation. For the moment, we assume that we can replace $y \to y^{\mu}$ and $\hat{y} \to \hat{y}^{\mu}$ directly in Eq.~\eqref{eq:cl1}, i.e. we fix the random variables on the observed data. Using the Dirac delta  as a representation of the Indicator function we have:
 \begin{equation}  \nonumber
 P(y) =  \int  \left[ \prod_{\mu} d \hat{y}^{\mu} \right]  \,  \prod_{\mu} \delta( y^{\mu} - \hat{y}^{\mu})  \, P(\hat{y})
\label{eqa:cl1}
 \end{equation}
 \begin{equation}
 =  \int  \left[ \prod_{\mu}  \frac{d\lambda^{\mu}}{2\pi} \right] \prod_{\mu}  e^{i \, \sum_{\mu} \lambda^{\mu} \, y^{\mu} } \chi(\lambda^{\mu}),
\end{equation}
where we have introduced $m$ Lagrange multipliers $\lambda$ to enforce the $\delta$-function constraint and identified  the characteristic function
\begin{align} \label{eqa:char}
\chi(\lambda^{\mu}) &=  \int  \left[ \prod_{\mu} d\hat{y}^{\mu} \right]  P(\hat{y}^{\mu} )  \, e^{-i \sum_{\mu} \lambda^{\mu} \, \hat{y}^{\mu}}
\end{align}
Let us consider the case of i.i.d. examples, distributed according to the Bernoulli distribution:
\begin{equation} \label{eqa:bern}
P(\hat{y}^{\mu} ) = \int d\boldsymbol{\theta} \left\{ q^{\mu}(\boldsymbol{\theta}) \, \delta(\hat{y}^{\mu}-1 ) + (1-q^{\mu}(\boldsymbol{\theta}) ) \, \delta(\hat{y}^{\mu}) \right\},
\end{equation}
where each outcome is either $1$ or $0$ with a {\it sample dependent} probability $q^{\mu}(\boldsymbol{\theta})$ being the output value of a Neural Network solving a binary classification task; as such, $q$ has the functional form of a {\it Sigmoid} function with $\boldsymbol{\theta}$ a set of network parameters. Eq.~\eqref{eqa:char} then reads
\begin{align} \label{eqa:char}
 \chi(\lambda^{\mu}) &=  \int  d\boldsymbol{\theta} \left[ q^{\mu}(\boldsymbol{\theta} ) e^{-i \, \lambda^{\mu} } +(1-q^{\mu}(\boldsymbol{\theta} )) \right].
 \end{align}
Using this expression back in Eq.~\eqref{eqa:bern} we arrive at the intermediate result
 \begin{align} \label{eqa:cl2} \nonumber
 P(y) &= \int d\boldsymbol{\theta} \left[ \prod_{\mu}  \frac{d\lambda^{\mu}}{2\pi} \right] e^{i \, \sum_{\mu} ( \lambda^{\mu} \, y^{\mu} + \log \left[ q^{\mu} e^{-i \, \lambda^{\mu} } +(1-q^{\mu}) \right] )}  \\
 &=  \int  \left[ \prod_{\mu}  \frac{d\lambda^{\mu}}{2\pi} \right] e^{m \, S[\lambda]}.
 \end{align}
For a large number of training examples $m$, we can solve the above integral using the steepest descent. This fixes the value of the Lagrange multipliers:
\begin{align} \label{eqa:cl3}
\frac{\partial S[\lambda]}{\partial \lambda^{\mu}} &= i \, y^{\mu} - \frac{i \, q^{\mu} \, e^{-i \, \lambda^{\mu}}}{q^{\mu} \, e^{-i \, \lambda^{\mu}} + (1-q^{\mu})} = 0 \\ \nonumber
&\rightarrow - i \lambda^{\mu}_{c} =  \log \frac{y^{\mu}(1-q^{\mu}) }{q^{\mu}(1- y^{\mu})}
\end{align}
Using the optima back in Eq.~\eqref{eqa:cl2} we arrive after some simple algebra to the expression for the cross-entropy:
\begin{align} \label{eq:cl3}
P(y) &\simeq \int d\boldsymbol{\theta} \, e^{m S[\boldsymbol{\theta}]} \\ \nonumber
S[\boldsymbol{\theta}] &= \frac{1}{m} \sum_{\mu} \left\{ y^{\mu} \, \log [ q^{\mu}( \boldsymbol{\theta} ) ] + (1-y^{\mu}) \, \log [1-q^{\mu}( \boldsymbol{\theta}) ] \right\},
\end{align}
up to an additive constant depending only on $y$. The final step consists of evaluating the $\theta$ integral again for $m \gg 1$, and take the maximum likelihood value $\theta^*$. This last step is numerically performed by the gradient descent algorithm. Note however that the evaluation of the cost function in the optima negelects ``finite size'' correcions, and it becomes more and more correct as the number of examples $m$ increases. In this derivation we assumed that the two outcomes of the network are not correlated, but one can relax this assumption and obtain a new Loss function that takes into account possible correlations among the labels. The latter case corresponds to multi-label classification where, controrary to multi-class classification, labels are mutually correlated.

Previously, we noticed that from a statistical mechanics point of view we are evaluating an {\it annealed} average; however, as the input variables are fixed this should rather be a {\it quenched} average. This is the case of spin glasses~\cite{parisi2, giardina}, where the quenched disorder (a fixed value of the couplings between spins) leads to frustration and therefore a highly non-convex energy landscape. Interestingly, a quenched scenario is similar to the approach proposed in~\cite{pratik}, leading to an improved optimization protocol. We leave this point as an open question for future research.

\bibliography{mfa}

\begin{thebibliography}{35}
\providecommand{\natexlab}[1]{#1}
\providecommand{\url}[1]{\texttt{#1}}
\expandafter\ifx\csname urlstyle\endcsname\relax
  \providecommand{\doi}[1]{doi: #1}\else
  \providecommand{\doi}{doi: \begingroup \urlstyle{rm}\Url}\fi

\bibitem[Amit(1989)]{amit1}
Amit, D.~J.
\newblock \emph{Modeling Brain Functions, The world of acctractor Neural
  Networks}.
\newblock Cambridge University Press, 1989.

\bibitem[Bellac et~al.(2004)Bellac, Mortessagne, and Bastrouni]{bellac}
Bellac, M.~L., Mortessagne, F., and Bastrouni, G.~G.
\newblock \emph{Equilibrium and Non-Equilibrium Statistical Thermodynamics}.
\newblock Cambridge University Press, 2004.

\bibitem[Bishop(2006)]{bishop}
Bishop, C.
\newblock \emph{Pattern recognition and machine learning}.
\newblock Springer, 2006.

\bibitem[Cassandro \& Jona-Lasinio(1978)Cassandro and Jona-Lasinio]{cassandro}
Cassandro, M. and Jona-Lasinio, G.
\newblock Critical point behaviour and probability theory.
\newblock \emph{Advances in Physics}, 27\penalty0 (6):\penalty0 913 -- 941,
  1978.

\bibitem[Chaudhari et~al.(2017)Chaudhari, Choromanska, Soatto, LeCun, Baldassi,
  Borgs, Chayes, Sagun, and Zecchina]{pratik}
Chaudhari, P., Choromanska, A., Soatto, S., LeCun, Y., Baldassi, C., Borgs, C.,
  Chayes, J., Sagun, L., and Zecchina, R.
\newblock Entropy-sgd: Biasing gradient descent into wide valleys.
\newblock \emph{arXiv:1611:01838}, 2017.

\bibitem[Choromanska et~al.(2015)Choromanska, Henaff, Mathieu, Michael, Gerard,
  and LeCun]{ch}
Choromanska, A., Henaff, M., Mathieu, M., Michael, A., Gerard, B., and LeCun,
  Y.
\newblock The loss surface of multilayer networks.
\newblock \emph{JMLR}, 38, 2015.

\bibitem[Dauphin et~al.(2014)Dauphin, Pascanu, Gulcehre, Cho, Ganguli, and
  Bengio.]{dauphin}
Dauphin, Y.~N., Pascanu, R., Gulcehre, C., Cho, K., Ganguli, S., and Bengio.,
  Y.
\newblock Identifying and attacking the saddle point problem in
  high-dimensional non-convex optimization.
\newblock \emph{Advances in Neural Information Processing Systems},
  27:\penalty0 2933 -- 2941, 2014.

\bibitem[Di-Castro \& Raimondi(2003)Di-Castro and Raimondi]{roberto}
Di-Castro, C. and Raimondi, R.
\newblock \emph{Statistical mechanics and applications in condensed matter}.
\newblock Cambridge University Press, 2003.

\bibitem[Dinh et~al.(2017)Dinh, Pascanu, Bengio, and Bengio]{Dinh2017}
Dinh, L., Pascanu, R., Bengio, S., and Bengio, J.
\newblock Sharp minima can generalize for deep nets.
\newblock \emph{Proceedings of the 34th International Conference on Machine
  Learning}, 2017.

\bibitem[Dominicis \& Giardina(2016)Dominicis and Giardina]{giardina}
Dominicis, C.~D. and Giardina, I.
\newblock \emph{Random Fields and Spin Glasees, A Field Theory approach}.
\newblock Cambridge University Press, 2016.

\bibitem[Elfwing et~al.(2014)Elfwing, Uchibe, and Doya]{elfwig}
Elfwing, S., Uchibe, E., and Doya, K.
\newblock Expected energy-based restricted boltzmann machine for
  classification.
\newblock \emph{Neural Networks}, 64:\penalty0 29--38, 2014.
\newblock \doi{http://dx.doi.org/10.1016/j.neunet.2014.09.006}.

\bibitem[Hayou et~al.(2018)Hayou, Doucet, and Rousseau]{soufiane}
Hayou, S., Doucet, A., and Rousseau, J.
\newblock On the selection of initialization and activation function for deep
  neural networks.
\newblock \emph{arXiv:1805.08266v1}, 2018.

\bibitem[Hertz et~al.(1991)Hertz, Krogh, and Palmer]{hertz}
Hertz, J., Krogh, A., and Palmer, R.~G.
\newblock \emph{Introduction to the Theory of Neural Computation}.
\newblock Introduction to the theory of Neural computation, 1991.

\bibitem[Jaynes(2003)]{jaynes}
Jaynes, E.~T.
\newblock \emph{Probability Theory, the logic of science}.
\newblock Cambridge University Press, 2003.

\bibitem[Kou et~al.(2018)Kou, Lee, and Ng]{connie}
Kou, C., Lee, H.~K., and Ng, T.~K.
\newblock Distribution regression networks.
\newblock \emph{arXiv:1409.6179v1}, 2018.

\bibitem[Krizhevsky et~al.(2012)Krizhevsky, Sutskever, and Hinton]{Krizhevsky}
Krizhevsky, A., Sutskever, I., and Hinton, G.~E.
\newblock Imagenet classification with deep convolutional neural networks.
\newblock \emph{Advances in neural information processing systems}, pp.\
  1097--1105, 2012.

\bibitem[Lin et~al.(2017)Lin, Tegmark, and Rolnick]{LinTegmark2017}
Lin, H.~W., Tegmark, M., and Rolnick, D.
\newblock Why does deep and cheap learning work so well?
\newblock \emph{Journal of Statistical Physics}, 168:\penalty0 1223--1247,
  2017.

\bibitem[Ma(1976)]{ma}
Ma, S.-K.
\newblock \emph{Modern Theory of Critical Phenomena}.
\newblock Advanced book program. Westview press, 1976.

\bibitem[MacKay(2003)]{mckay}
MacKay, D.~J.
\newblock \emph{Information Theory, Inference and Learning algorithms}.
\newblock Cambridge University Press, 2003.

\bibitem[Mehta \& Schwab(2014)Mehta and Schwab]{mehta}
Mehta, P. and Schwab, D.~J.
\newblock An exact mapping between the variational renormalization group and
  deep learning.
\newblock \emph{arXiv:1410.3831}, 2014.

\bibitem[M\'ezard(2018)]{mezard1}
M\'ezard, M.
\newblock Artificial intelligence and its limits.
\newblock \emph{Europhysics News}, 49, 2018.

\bibitem[M\'ezard \& Montanari(2009)M\'ezard and Montanari]{mezard}
M\'ezard, M. and Montanari, A.
\newblock \emph{Information, Physics and Computation}.
\newblock Oxford University Press, 2009.

\bibitem[M\'ezard et~al.(1987)M\'ezard, Parisi, and Virasoro]{parisi2}
M\'ezard, M., Parisi, G., and Virasoro, M.~A.
\newblock \emph{Spin Glass Theory and Beyond}, volume~9 of \emph{Lecture notes
  in Physics}.
\newblock World Scientific, 1987.

\bibitem[Nguyen et~al.(2017)Nguyen, Zecchina, and Berg]{zecchina}
Nguyen, H.~C., Zecchina, R., and Berg, J.
\newblock Inverse statistical problems: from the inverse ising problem to data
  science.
\newblock \emph{Advances in Physics}, pp.\  197--261, 2017.

\bibitem[Pennington \& Bahri(2017)Pennington and Bahri]{penn1}
Pennington, J. and Bahri, Y.
\newblock Geometry of neural network loss surfaces via random matrix theory.
\newblock \emph{Proceedings of the 34th International Conference on Machine
  Learning, Sydney, Australia}, 2017.

\bibitem[Pennington \& Worah(2017)Pennington and Worah]{penn2}
Pennington, J. and Worah, P.
\newblock Nonlinear random matrix theory for deep learning.
\newblock \emph{31st Conference on Neural Information Processing Systems},
  2017.

\bibitem[Pogio et~al.(2017)Pogio, Mhaskar, Rosasco, Miranda, and
  Liao.]{Poggio2017}
Pogio, T., Mhaskar, H., Rosasco, L., Miranda, B., and Liao., Q.
\newblock Why and when can deep-but not shallow-networks avoid the curse of
  dimensionality: A review.
\newblock \emph{International Journal of Automation and Computing},
  14:\penalty0 503--519, 2017.

\bibitem[Poole et~al.(2016)Poole, Lahiri, Raghu, Sohl-Dickstein, and
  Ganguli]{Poole2016}
Poole, B., Lahiri, S., Raghu, M., Sohl-Dickstein, J., and Ganguli, S.
\newblock Exponential expressivity in deep neural networks through transient
  chaos.
\newblock \emph{Advances in Neural Information Processing Systems},
  29:\penalty0 3360 -- 3368, 2016.

\bibitem[Raghu et~al.(2016)Raghu, Poole, Kleinberg, Ganguli, and
  Sohl-Dickstein.]{Raghu2017}
Raghu, M., Poole, B., Kleinberg, J., Ganguli, S., and Sohl-Dickstein., J.
\newblock On the expressive power of deep neural networks.
\newblock \emph{Proceedings of the 34th International Conference on Machine
  Learning}, 1\penalty0 (2847-2854), 2016.

\bibitem[Ramachandran et~al.(2017)Ramachandran, Zoph, and Le]{prajit}
Ramachandran, P., Zoph, B., and Le, Q.~V.
\newblock Searching for activation functions.
\newblock \emph{arXiv:1710.05941}, 2017.

\bibitem[Sagun et~al.(2017)Sagun, Bottou, and LeCun.]{levent}
Sagun, L., Bottou, L., and LeCun., Y.
\newblock Eigenvalues of the hessian in deep learning: Singularity and beyond.
\newblock \emph{arXiv:1611.07476v2}, 2017.

\bibitem[Schwartz-Ziv \& Tishby(2017)Schwartz-Ziv and Tishby]{tishby2}
Schwartz-Ziv, R. and Tishby, N.
\newblock Opening the black box of deep neural networks via information.
\newblock \emph{arXiv:1703.00810}, 2017.

\bibitem[Srivastava et~al.(2014)Srivastava, Hinton, Krizhevsky, Sutskever, and
  Salakhutdinov]{srivastava2014}
Srivastava, N., Hinton, G., Krizhevsky, A., Sutskever, I., and Salakhutdinov,
  R.
\newblock Dropout: A simple way to prevent neural networks from overfitting.
\newblock \emph{Journal of Machine Learning Research}, 15:\penalty0 1929 --
  1958, 2014.

\bibitem[Tishby \& Zaslavsky(2015)Tishby and Zaslavsky]{tishby1}
Tishby, N. and Zaslavsky, N.
\newblock Deep learning and the information bottleneck principle.
\newblock \emph{Invited paper to ITW 2015; 2015 IEEE Information Theory
  Workshop}, 2015.

\bibitem[Zhang et~al.(2017)Zhang, Bengio, Hardt, Recht, and Vinyals]{Zhang2017}
Zhang, C., Bengio, S., Hardt, M., Recht, B., and Vinyals, O.
\newblock Understanding deep learning requires rethinking generalization.
\newblock \emph{arXiv:1611.03530}, 2017.

\end{thebibliography}
\bibliographystyle{icml2019}

\end{document}